\documentclass[letters,10pt]{fcs}

\usepackage{graphics} 
\usepackage{epsfig} 
\usepackage{mathptmx} 
\usepackage{times} 
\usepackage{amsmath} 
\usepackage{amssymb} 
\usepackage{xcolor}
\usepackage{booktabs}
\usepackage{graphicx}
\usepackage{multirow}
\usepackage{listings}
\usepackage{url}
\usepackage[utf8]{inputenc}
\usepackage{amsfonts}
\usepackage{algorithm}
\usepackage{algpseudocode}
\usepackage{hyperref}
\usepackage{float}
\definecolor{mygreen}{RGB}{109,204,109}
\definecolor{myblue}{RGB}{85,172,213}
\definecolor{myyellow}{RGB}{255,223,96}
\definecolor{myred}{RGB}{192,0,0}
\usepackage{footnote}
\usepackage{microtype}

\title{BestMan: A Modular Mobile Manipulator Platform \\for Embodied AI with Unified Simulation-Hardware APIs}

\shorttitle{BestMan: A Mobile Manipulator Platform}

\newcommand{\equalcontribution}{\textsuperscript{†}}
\newcommand{\correspondingauthors}{\textsuperscript{*}}

\author[1]{Kui Yang\equalcontribution}
\author[2]{Nieqing Cao\equalcontribution}
\author*[3]{Yan Ding\correspondingauthors}
\author*[1]{Chao Chen\correspondingauthors}
\address[1]{College of Computer Science, Chongqing University}
\address[2]{Xi'an Jiaotong-Liverpool University}
\address[3]{Shanghai Artificial Intelligence Laboratory}
\fcssetup{
  corr-email     = {yding25@binghamton.edu, cschaochen@cqu.edu.cn },
}

\begin{document}

\renewcommand{\thefootnote}{\fnsymbol{footnote}}
\footnotetext[2]{Equal contribution, * Corresponding author}

\section{Introduction}
Embodied Artificial Intelligence (Embodied AI) has recently become a key research focus~\cite{duan2022survey}.
It emphasizes agents' abilities to perceive, comprehend, and act in physical worlds to complete tasks. 
\emph{Simulation platforms} are essential in this area, as they simulate agent behaviors in set environments and tasks, thereby accelerating algorithm validation and optimization.
However, constructing such a platform presents several challenges.

\emph{One primary challenge is the complexity of multilevel technical integration}.
Completing tasks generally involves multiple tightly interconnected layers, including perception, planning, and control, each requiring specialized knowledge and algorithms~\cite{zhao2024survey}.
These layers' interdependencies require precise coordination, making platform development challenging.
Building upon this, \emph{the insufficient modularity in existing platforms further limits expandability and algorithm integration.}
Existing platforms may be modular at a high level but often lack clear internal modularity in key components.
For instance, though Habitat platform~\cite{habitat19iccv} offers broad functionality, its complex setup and dependency management reduce plug-and-play usability.
Its tightly coupled architecture and monolithic code make it hard to replace or extend key modules like manipulation.
\emph{Moreover, interface heterogeneity between simulation environments and physical robotic systems impedes algorithm deployment.} 
Algorithms validated in simulation are often not directly transferable to hardware platforms.
This incompatibility arises from differences in hardware interfaces, device drivers, and system dependencies between virtual and real-world systems. 
Consequently, significant effort is required to adapt or reimplement algorithms for hardware execution, increasing development time and complexity.
\emph{Finally, adapting to diverse mobile manipulators presents substantial obstacles.} 
Mobile manipulators vary widely, encompassing different mobile bases, robotic arms, and end-effectors. 
A platform that adapts to various mobile manipulators, decouples software from hardware, and reduces development challenges is urgently needed.

\begin{figure*}
    \centering
    \includegraphics[width=0.95\linewidth]{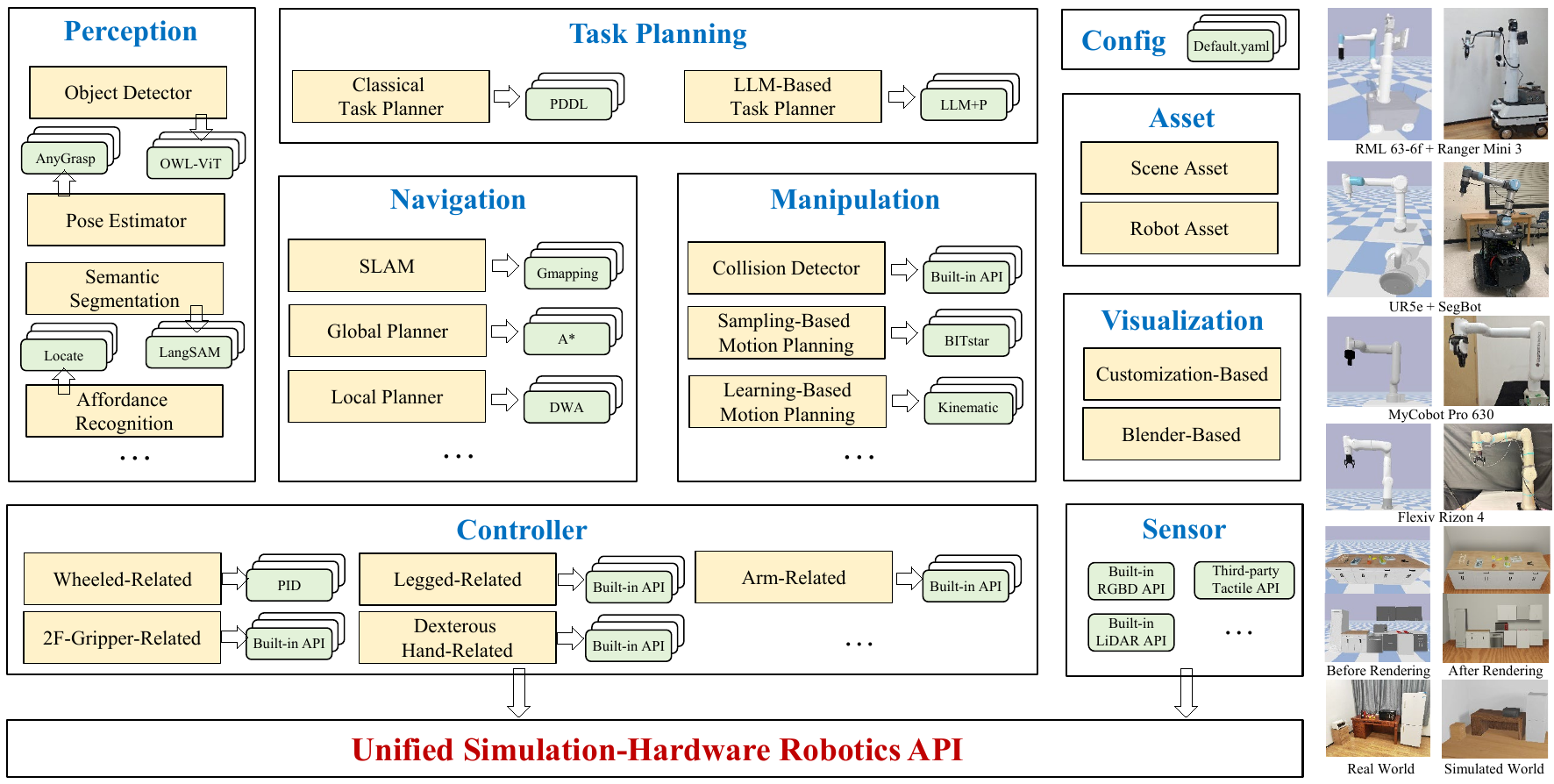}
    \caption{\textbf{Overview of Platform Architecture.}
    The platform comprises ten major components (highlighted in \textcolor{myblue}{blue} and \textcolor{myred}{red}): \emph{Perception, Task Planning, Navigation, Manipulation, Configuration, Asset, Visualization, Controller, Sensor}, and \emph{Robotics API}. 
    Each component contains modules (highlighted in \textcolor{myyellow}{yellow}), where various algorithms can be implemented, represented by rounded rectangles, with the default methods highlighted in \textcolor{mygreen}{green}. 
    The ellipsis (`$\cdots$') indicates customizable modules or algorithms that users can extend. 
    The unified simulation-hardware robotics APIs are constructed based on the control and sensing components, while other components are independent of these robotics APIs. 
    The right panel illustrates the platform's applicability across various real and simulated mobile manipulators and environments.
    }
    \label{fig:architecture}
\end{figure*}

\renewcommand{\thefootnote}{\arabic{footnote}}
To address these challenges, we develop the \textbf{BestMan}\footnote{Project Website: \url{https://github.com/AutonoBot-Lab/BestMan_Pybullet}} platform based on the PyBullet simulator~\cite{pybullet}, with the following key contributions:
1) \emph{Integrated Multilevel Skill Chain to Address Multilevel Technical Complexity}. 
We provide an integrated simulation platform covering multiple layers, including perception, task planning, motion planning, and control.
By providing a skill chain, we streamline coordination between these layers, thereby reducing the difficulty of integrating different components and enhancing overall development efficiency.
2) \emph{Highly Modular Design for Expandability and Algorithm Integration}.
We design skill modules with standardized interfaces that support arbitrary internal algorithm composition, ensuring modularity within key components. 
This allows users to easily replace or extend algorithms, reducing development complexity. 
3) \emph{Unified Interfaces for Simulation and Real Devices to Address Interface Heterogeneity}: 
We implement unified API interfaces for both simulation and real devices. 
In skill modules, the interfaces maintain consistency in naming and functionality, differing only in their underlying implementations. 
This enables algorithms developed and tested in the simulation to be efficiently migrated to real devices, thus addressing the challenge of simulation-to-reality discrepancies, and reducing adaptation and reimplementation efforts.
4) \emph{Decoupling Software from Hardware to Address Hardware Diversity}:
We decompose mobile manipulators into modular components, including mobile bases, robotic arms, and grippers, allowing flexible combinations and replacements of hardware. 
This modularity simplifies adapting to different hardware configurations and reduces development effort.

\vspace{-0.1em}
\section{Platform Architecture}
Fig.~\ref{fig:architecture} presents the architectural overview of the BestMan platform with ten key components.
These components are designed to address key research areas in Embodied AI.
Details are in the appendix\footnote{\url{https://github.com/AutonoBot-Lab/BestMan_Pybullet/blob/master/docs/appendix}}.

The \emph{Perception} component processes sensor data, particularly RGB-D information to generate various predictions,  including affordance recognition.
For example, in an object interaction task, such as opening a fridge, the handle would be identified as the key affordance.
Based on the perception outcomes, \emph{Task Planning} generates the action sequence to complete the service task.
The platform supports two primary groups of task planning approaches, corresponding to established strategies in robotics~\cite{zhao2024survey}.
The resulting action sequence is divided into navigation and manipulation actions.
Accordingly, two major components are provided: \emph{Navigation} and \emph{Manipulation}. 
The Navigation component includes modules for SLAM, global planning, and local planning. 
The \emph{Manipulation} component implements two trajectory computation approaches: sampling-based and learning-based motion planning.
For action execution in the simulation environment, the \emph{Controller} component mainly uses PyBullet's built-in control APIs.
Controllers are categorized by robot type, including categories for wheeled, legged, and arm-based robots.
The platform also includes an \emph{Asset} component, further divided into \emph{Scene Assets} and \emph{Robot Assets}, allowing flexible configuration of both environments and robot models for various tasks.
Given PyBullet's limitations in visualization, a Blender plugin enhances PyBullet's visualization, making it more suitable for demonstration purposes.
The \emph{Sensor} component extends beyond built-in RGB-D sensors, incorporating third-party tactile sensors, which are particularly useful for tasks requiring precision in contact and force feedback.
Finally, the \emph{Configuration} component uses YAML files to manage hyperparameters, improving reusability and simplifying the tuning process.

\section{Platform Highlights}
The BestMan platform is designed to address the key challenges through a set of deliberate design choices and technical solutions. 
Below, we outline how the platform addresses each of these challenges.
BestMan's \emph{integrated multilevel skill chain} coordinates essential layers such as perception, task planning, navigation, manipulation, and control, providing the necessary capabilities for robots to perform a wide range of service tasks, particularly in home environments.

The \emph{platform's modularity} is evident in its hierarchical structure.
At the high level, the platform is divided into key components.
Each component is further subdivided into \emph{modules}, allowing users to add or modify modules to customize and extend the platform's capabilities.
Within each module, algorithms are encapsulated as independent units, facilitating the straightforward integration or replacement of algorithms.
For instance, in the Pose Estimator module, the default algorithm (AnyGrasp~\cite{fang2023anygrasp}) can be replaced with a custom implementation. 
Each module is organized as a subfolder within its respective component, with individual algorithms packaged as separate Python classes. 
This organization ensures that each algorithm operates independently, enabling flexible expansion and customization.
\emph{Only the Controller and Sensor components interface directly with the unified Robotics API, while all other components remain decoupled.} 
This separation allows custom algorithms without dependencies on the API, streamlining integration and improving scalability.

Another key challenge is bridging \emph{the gap between simulation and real-world systems.}
Simulated environments like PyBullet provide built-in APIs for convenient control in virtual spaces.
However, these simulation-specific controllers are not directly applicable to real-world robots due to differences in hardware interfaces, device drivers, and control mechanisms across platforms.
To bridge this gap, the BestMan platform introduces a \emph{middleware abstraction layer} offering unified APIs for both simulated and real devices.
This Unified Simulation-Hardware Robotics API decouples hardware- and software-specific implementations, enabling algorithms validated in simulation to transfer to real robots with minimal modification.
The API interface remains consistent—e.g., a function like \texttt{move\_forward()} exists in both environments—but the underlying implementation differs.
In PyBullet, \texttt{move\_forward()} might use a PID controller, whereas on a real robot, it invokes the manufacturer's custom motor control API.
Despite these differences, the unified API allows users to apply the same high-level commands in both environments without concern for hardware-specific details.
This approach is crucial for the \emph{Controller} and \emph{Sensor} components, which are responsible for executing actions and receiving feedback from both real and simulated environments. 

The platform's modular hardware architecture offers flexibility in \emph{adapting to a variety of mobile manipulators}. 
The \emph{Asset} component includes \emph{Scene Assets} (e.g., URDF models of common objects) and \emph{Robot Assets}, which consist of modular units for \emph{wheels, legs, arms, grippers}, and \emph{hands}.
These units can be combined to configure different robotic systems, facilitating adaptation to diverse hardware setups. 
By decoupling software from hardware, the platform enables experimentation with various combinations of mobile bases, robotic arms, and end-effectors without requiring changes to the underlying software, thereby reducing development complexity and time.


\bibliographystyle{fcs}
\bibliography{ref}

\begin{thebibliography}{1}

\bibitem{duan2022survey}
Duan J, Yu~S, Tan H~L, Zhu H, Tan C.
\newblock A survey of embodied {AI}: From simulators to research tasks.
\newblock IEEE Transactions on Emerging Topics in Computational Intelligence, 2022, 6(2): 230--244

\bibitem{zhao2024survey}
Zhao Z, Cheng S, Ding Y, Zhou Z, Zhang S, Xu~D, Zhao Y.
\newblock A survey of optimization-based task and motion planning: From classical to learning approaches.
\newblock IEEE/ASME Transactions on Mechatronics, 2024

\bibitem{habitat19iccv}
{Manolis Savva*} , {Abhishek Kadian*} , {Oleksandr Maksymets*} , Zhao Y, Wijmans E, Jain B, Straub J, Liu J, Koltun V, Malik J, Parikh D, Batra D.
\newblock Habitat: {A} platform for embodied {AI} research.
\newblock In: Proceedings of the IEEE/CVF International Conference on Computer Vision (ICCV).
\newblock 2019

\bibitem{pybullet}
Coumans E, Bai Y.
\newblock Pybullet, a python module for physics simulation for games, robotics and machine learning.
\newblock \url{http://pybullet.org}, 2016--2022

\bibitem{fang2023anygrasp}
Fang H~S, Wang C, Fang H, Gou M, Liu J, Yan H, Liu W, Xie Y, Lu~C.
\newblock Anygrasp: Robust and efficient grasp perception in spatial and temporal domains.
\newblock IEEE Transactions on Robotics, 2023

\end{thebibliography}


\end{document}